\documentclass[journal]{IEEEtran}
\usepackage{graphicx}
\usepackage{gensymb}
\usepackage{hyperref}
\usepackage{amsmath}
\usepackage{amssymb}
\usepackage{cite}
\IEEEoverridecommandlockouts  
\title{\LARGE \bf Simulating Robotic Locomotion in Sand: Resistive Force Theory in an Open-Source Physics Engine}
\author{{Ryan Walker Brown, Laura K. Treers, Kathryn A. Daltorio},~\IEEEmembership{Member,~IEEE}

\thanks{Currently under review at IEEE Transactions in Robotics}
\thanks{This work was supported by the Office of Naval Research under the grant ONR N00014-24-1-2022}
\thanks{Corresponding Author Ryan Walker Brown: ryanbrown@case.edu, and Kathryn A. Daltorio: kati@case.edu}
\thanks{Ryan W. Brown and Kathryn A. Daltorio are with the College of Engineering, Department of Mechanical and Aerospace Engineering at Case Western Reserve University, Cleveland, OH 44106}
 \thanks{Laura K. Treers is with the College of Engineering and Mathematical Sciences, Department of Mechanical Engineering at the University of Vermont, Burlington, VT 05405
}}
\begin{document}
\maketitle

\begin{abstract}
Recent advancements in Resistive Force Theory (RFT) enable approximation of ground reaction forces for locomotion in sand without the computational expense of modeling interactions with individual grains. However, these tools have been absent in 3D physics engines commonly used for robot simulation. We explore if resistive force approximations are sufficient, when integrated with standard dynamics calculations, to provide a stable substrate for a freely walking robot. To determine this, we implement 3D Granular Resistive Force Theory (3D RFT) in a physics simulation engine, MuJoCo.  We verify simulations in multiple scenarios to demonstrate that key trends due to end effector shape, speed, and loading are preserved. Our implementation predicts both walking distance and foot sinkage of a 12-Degree of Freedom hexapod robot within 7\% of experiments in sand. 
While RFT has inherent approximations, the open source tool described here has potential to help develop new and improved robot designs to traverse granular media substrates.  
\end{abstract}

\vspace{-4mm}
\section{Introduction}

The unique behavior of granular media is so universal that “sandbox” has become shorthand for “a controlled virtual environment for creative development” \cite{devanga_creating_2025,ringe_regulating_2020,blasing_android_2010}
 -- and yet no widely-available robotics simulation software models sand. 
In man-made environments, factory robots are optimized using simulated digital twins to streamline supply chains \cite{becue_new_2020, zhang_modeling_2019}, cars drive more reliably with simulation-developed collision prevention \cite{karunakaran_efficient_2020} or gridlock prevention \cite{sivakumar_design_2023}, and legged robots are learning in simulation to open doors \cite{arm_pedipulate_2024} and climb stairs \cite{qi_perceptive_2021}.  In outdoor environments, sinking into sand affects step size, grain size affects cost of transport, and dirt refilling holes after a step affects foot placement. Excluding these common effects from simulation limits the complexity and robustness of behaviors that are possible for outdoor robots. By moving from linear stiffness models to depth-dependent resistive forces in an open-source simulator, we hope to provide a “sandbox" for robots that traverse new paths in natural environments.  

Currently, the most developed robots for granular terrains are either heavy duty, heavily sensorized, or tested in specialized physical environments. For example, heavy construction vehicles\cite{scora_variability_2019}, agricultural tractors\cite{molari_performance_2012}, and all-terrain tanks\cite{kim_dominant_2021} are scaled such that the sinkage into the substrate is relatively small. NASA and other space agencies have identified mars-like deserts for testing\cite{azua-bustos_atacama_2022} and constructed testbeds for rovers \cite{rezich_nasa_2021}.  With many sensors, smart bipeds \cite{gosyne_bipedial_2018} and quadrupeds \cite{kolvenbach_traversing_2022} have demonstrated walking in sand based on closed-loop reactions \cite{zheng_intruder_2018}. 

 \begin{figure}[t]
    \centering
    \includegraphics[width=.65\columnwidth]{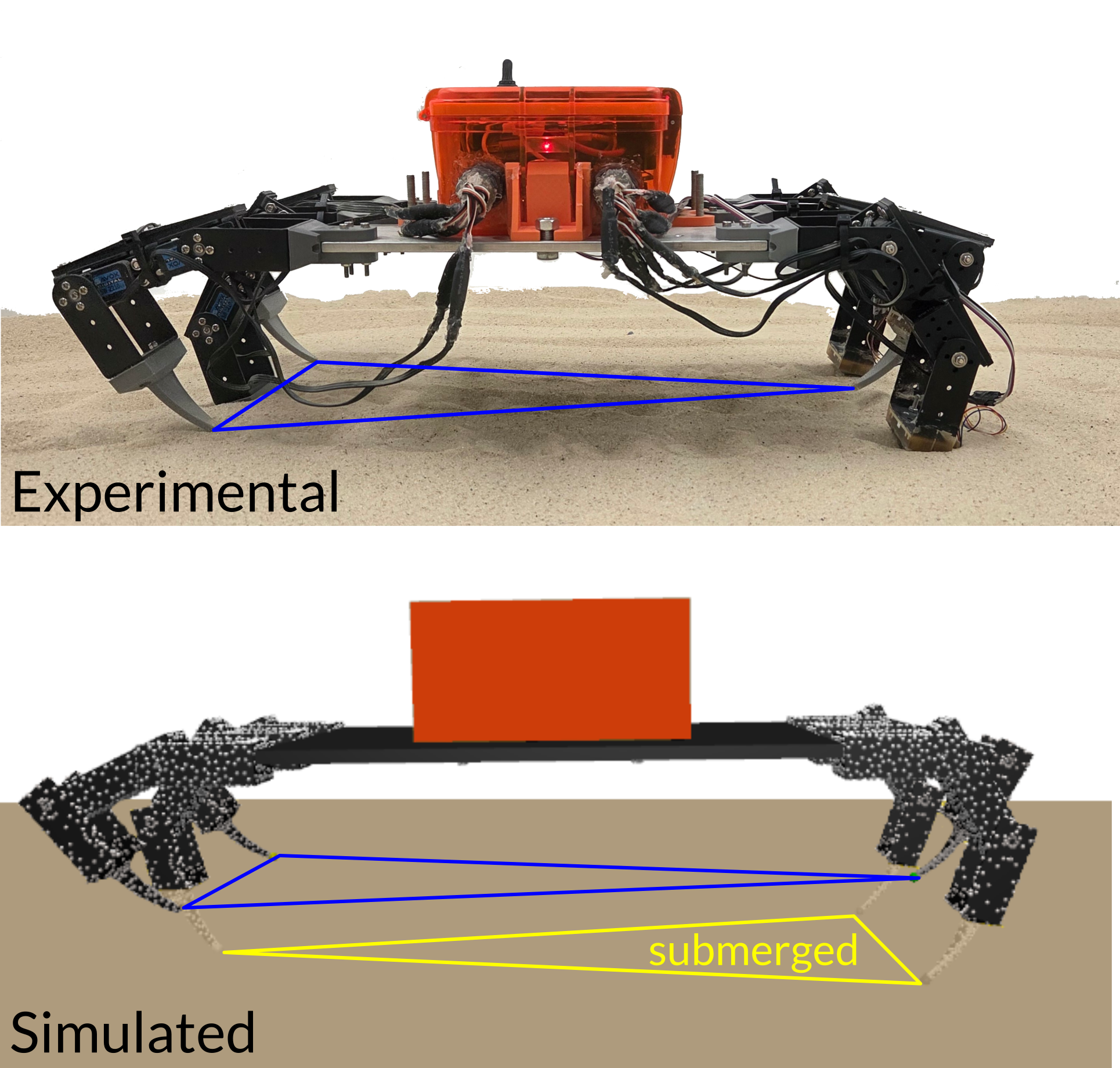}  %
    \caption{\textbf{Walking of legged robots in granular media is modeled using MuJoCo.}
       As a physical robot walks through sand (top), the legs and feet penetrate the ground at different angles throughout the gait cycle, especially when taking advantage of crab-like pointed end effectors (dactyls). To simulate (bottom), we use open-source MuJoCo for overall dynamics and add Resistive Force Theory (RFT) sites (light gray dots) across the leg surfaces.  As a result, the simulated stance dactyls (yellow tripod) sink into the ground until resistive forces support weight and inertial forces. The sinkage affects performance, in particular, determining the step size of swing dactyls (blue tripod).}
        \vspace{-5mm}
    \label{fig:realworld}
\end{figure}

\begin{figure*}[t!]
    \centering
    \includegraphics[width = .8\textwidth]{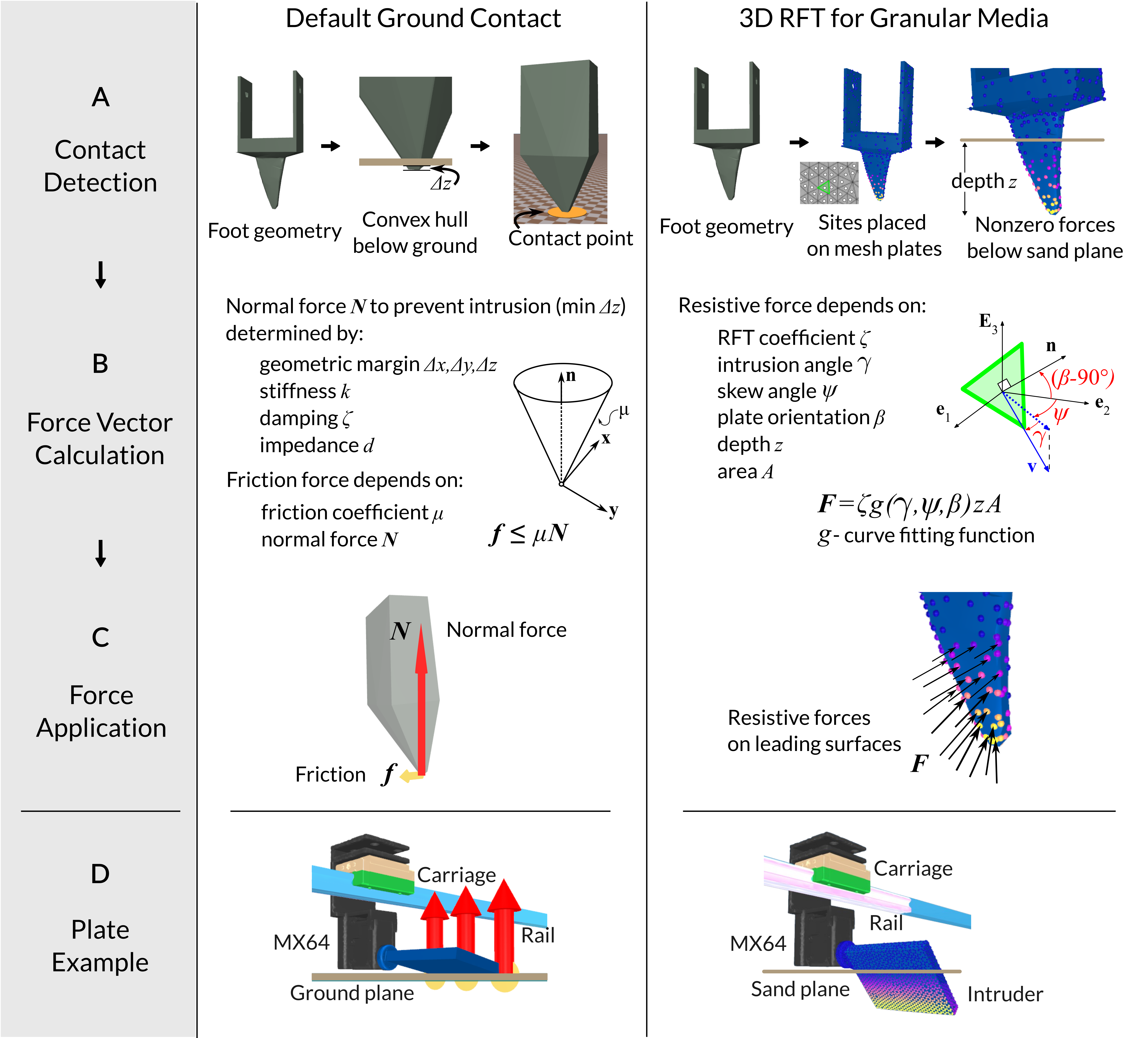}  %
    \caption{\textbf{Resistive Force Theory, RFT, is a fundamentally different approach for predicting ground contact forces, which estimates intrusion forces on rigid bodies passing through sandy ground.}
    (Left) MuJoCo's fast default handling of many contacts can be attributed to key simplifications of Coulomb friction. (A) First, the geometry is simplified and contact points where rigid convex hulls intersect are identified. (B) Rather than directly calculate the force required to prevent intersection, normal forces are approximated such that they increase with each timestep as constraints are violated. 
    (C) These normal forces and proportional friction forces are applied at the contact point. (D) The default approach does not allow legs to pass through ground. If an intruder rotates around a point with fixed height, the normal forces (which MuJoCo shows as three red arrows) increase indefinitely to prevent penetration. 
    (Right) Our RFT implementation integrates with MuJoCo by (A) discretizing geometry into plate elements and placing sites on each plate surface centroid, (B) determining intrusion force vectors based on plate orientation $\beta$, intrusion angle $\gamma$, depth \textit{z}, and area \textit{A}, and
   (C) applying resistive forces at each plate center. As as a result, (D) the RFT-affected intruder intrudes deep into the sand.  
   }    
    \label{fig:combo}
    \vspace{-5mm}
\end{figure*}
However, smaller, simpler robots \cite{treers_mole_2022, naclerio_controlling_2021, isava_experimental_2016, gong_using_2025, bagheri_bio-inspired_2024, tao_sbor_2020, russell_crabot_2011, barenboim_steerable_2020, zhang_mole-inspired_2024} may not have access to the same level of sensory feedback or precisely calibrated test beds  --- yet still sink considerably into sand such that sand-robot interaction forces cannot be neglected. Often these designs are inspired by animals that dig \cite{faulkes_coordination_1997}, swim \cite{maladen_biophysically_2010}, and undulate \cite{filardo_vehicle_2023} in sandy environments. To iteratively improve these robot behaviors, roboticists make in situ modifications to end-effectors, gait parameters, or payloads \cite{graf_addressing_2024}.  However, in situ modifications are not always possible or convenient. 

To capture the interaction forces when robot limbs or wheels sink into soft ground (Figure~\ref{fig:realworld}), some robot designers apply terramechanics principles to estimate nonlinear ground reaction forces. The most accurate approach, Discrete Element Method (DEM) \cite{dosta_comparing_2024}  computes motions of each individual grain, but the high computational cost is a limitation. Because many iterative design algorithms \cite{tan_comprehensive_2021}, learning methods \cite{karoly_deep_2021}, and model predictive control strategies \cite{salzmann_real-time_2023} often depend on simulations that run faster than real time, reduced-order empirical \cite{zhang_effectiveness_2014} and continuum approaches \cite{dunatunga_continuum_2015} are attractive. 

Resistive Force Theory (RFT) \cite{zhang_effectiveness_2014} is efficient and simple to use for estimating granular intrusion forces. RFT was originally developed for low Reynolds number fluids and was later extended to grains \cite{gray_propulsion_1955}.
Unlike typical friction-cone contact models\cite{kao_contact_2016} in which maximum tangential forces scale with normal forces, RFT forces in all directions are proportional to penetration depth, $z$, and surface area, $A$ (Figure~\ref{fig:combo}). Scaling factors are determined based on the direction of motion $\gamma$ and the angle of the surface $\beta$ \cite{li_terradynamics_2013}. Impressively, a single constant (the resistive coefficient, $\zeta$) can be used to calibrate different soils across a wide range of grain sizes and densities. Furthermore, RFT assumes that forces on individual discretized elements are independent, and thus are summed to estimate whole-body intrusion forces. The simplicity of this method enables RFT to provide rapid intrusion force predictions with reasonable accuracy, despite not directly modeling individual grains within the substrate.

Ongoing research has evaluated and expanded the accuracy of RFT for application to a range of locomotion behaviors, from rigid wheels \cite{agarwal_modeling_2019, yu_modeling_2024}, running hexapods \cite{li_terradynamics_2013}, and undulation \cite{li_compliant_2021}. RFT has also recently been expanded from the vertical \cite{li_terradynamics_2013} and horizontal \cite{maladen_undulatory_2009} planes to 3D \cite{treers_granular_2021, agarwal_mechanistic_2023} via the introduction of an additional characteristic angle, $\psi$.  Several implementations of 3D-RFT have been developed in recent years using these new kinematic parameters; namely, Treers et al. \cite{treers_granular_2021}, and Huang et al. \cite{huang_dynamic_2022} developed 3D implementations which combined elements of existing 2D formulations and Agarwal et al. \cite{agarwal_mechanistic_2023} developed a 3D implementation using a mechanistic framework.  For higher speeds, implementation of an inertial term is used in Dynamic RFT (D-RFT), but inertial effects were limited (speed \textless 20~cm/s)
in our simulations and thus we did not implement an inertial correction \cite{huang_dynamic_2022}. The implementation of Treers et al.'s formulation proved to be compatible with the MuJoCo framework because it was developed open-source as a standalone function that can determine forces on any meshed input geometry. 

RFT has known limitations, such as the inability to model granular jamming and shadowing \cite{ko_shadow_2000}, and the need for additional adjustments for cohesive soils \cite{kerimoglu_extending_2025}. However, RFT remains the most rapid method for estimating forces in granular media. The force approximations are surprisingly effective, even for burrowing in terrain. For example, optimizing the relative leg orientation to maximize downward force enabled a mole crab-inspired robot to dig into sand until its entire body was submerged \cite{treers_mole_2022}. However, solving the inverse problem (predicting motions from estimated forces) is more complicated due to the sensitivity of RFT at low velocity. RFT relies on a velocity vector from which to compute intrusion angle and thus intrusion forces; furthermore, when velocity magnitude approaches zero, the velocity direction used for RFT computation is ill-defined. 

Previous RFT implementations have predicted the motion of robots with continuously rotating legs \cite{li_terradynamics_2013}. In limited 2D scenarios such as for mini-RHex with MBDyn \cite{li_terradynamics_2013} and for multi-link fins with Pynamics \cite{li_compliant_2021}, motion was predicted, but none of the 3D extensions of RFT were used. Whether RFT is appropriate for predicting motion of multi-jointed legs, where joint velocities can result in a stationary stance, is untested. Here, we hypothesize that relatively simple smoothing will enable RFT to extend to articulated leg movement where ankle and knee joints minimize end effector velocity during stance. To our knowledge, 3D RFT has never been implemented in a simple open-source dynamic solver. In particular, in the vertical direction, a dynamic solver should integrate forces from a substrate and gravity to determine sinkage, which affects traction, efficiency, and chassis clearance.

Here, we investigate the extent to which a 3D RFT model, integrated into a multi-body simulator, approximates robot locomotion speed and ground reaction forces on sandy terrain. We select MuJoCo as the open-source dynamic solver to augment  because it is widely used \cite{erez_simulation_2015, blanco-mulero_benchmarking_2024}, and MuJoCo provides a faster and more capable environment than others such as Bullet, PhysX, and ODE \cite{erez_simulation_2015}.
We call the resulting implementation RFT-SiM (Resistive Force Theory --- Sand in MuJoCo) and compare prediction  
with physical experiments in a simple sandbox (Figure~\ref{fig:realworld}). Compared to previous implementations, RFT-SiM increases usability by enabling field simulations to run efficiently on a consumer-grade computer. 

For increasingly complex scenarios leading up to a freely walking 12-DOF robot, we show that varying leg speed, foot shape, and payload in the model correctly predicts changes in robot motion.  By predicting motion, rather than forces, our open-source approach, RFT-SiM, will enable roboticists to apply end-to-end learning algorithms and broader iterative mechanical design. This learning will power newer, cheaper (both monetarily and computationally) platforms in legged robotics. Furthermore, demonstrating  the ease of integration will encourage refinements and calibrations of terramechanics models to help robots navigate new areas - undersea, extraterrestrial, agriculture, and construction. 
Granular media simulations are also key tools to understand animals who thrive on different terrains, which helps us appreciate and mimic the creativity of the natural world.


The main contributions of this work are: (1) Development of an open-source granular media solver for MuJoCo using 3D RFT.
(2) Demonstration of algorithmic modification (force smoothing) and force application function for RFT. (3) Four physical characterization experiments from 1-DOF to 12-DOF which are compared with RFT-SiM. (4) Comparison of RFT-SiM accuracy with built-in MuJoCo contact algorithms for a hexapod robot walking in sand.

\vspace{-4mm}
\section{Methods}
We select a crab-like robot platform as an example model of animal-inspired locomotion in granular media. Legs of biological crabs tend to penetrate deeply into the substrate to provide stability without adding weight \cite{graf_crab-like_2019}. Crabs migrate across diverse granular media and serve as inspiration for the growing interest in legged all-terrain platforms. In addition, unlike other robots evaluated in sand \cite{li_terradynamics_2013, treers_mole_2022}, crab-like walking involves low velocity and stall of legs due to their pointed feet and varied gaits, both of which cause high sinkage (Figure~\ref{fig:realworld})\cite{grezmak_terrain_2021}. These conditions cannot be modeled with existing simulation tools; thus, we develop RFT-SiM as the first open-source framework appropriate for articulated walking when feet sink into sand.

\vspace{-5mm}
\subsection{RFT Implementation} \label{RFT Implementation in MuJoCo}
RFT-SiM replaces MuJoCo's built-in ground contact force calculation with RFT force calculations. This replacement required a few key changes to the standard MuJoCo physics (Figure~\ref{fig:combo}A-C).  First, each body is represented as a mesh of ``plates" to correctly capture the orientations of each surface and allow for RFT superposition of forces. Points (MuJoCo sites) are placed at the centroid of each mesh plate to serve as force application locations. Second, the orientation of the plates and their velocity vectors are used to calculate a set of force vectors based on the 3D RFT formulation described by Treers et al. \cite{treers_granular_2021}.

At low speeds (\textless 20~cm/s), RFT forces are independent of intrusion velocity magnitude and depend only on surface orientation, surface area, depth, and velocity direction. The interaction forces from RFT scale linearly with depth and cross-sectional area of a plate. Each plate of a mesh has its own local basis ($\textbf{e}_{1_j}$, $\textbf{e}_{2_j}$, and $\textbf{e}_{3_j}$), and the world frame is represented by $\textbf{E}_1$, $\textbf{E}_2$, and $\textbf{E}_3$ (Figure~\ref{fig:combo}B). $\textbf{e}_{1_j}$ represents the intersection of the plate with the horizontal plane $\textbf{E}_1$-$\textbf{E}_2$ while $\textbf{e}_{2_j}$ represents the horizontal projection of the plate's normal vector.

The 3D force, $\mathbf{F}_j$, from granular media is defined for a plate element \textit{j} as
\begin{align}
\mathbf{F}_j &= \zeta(\mathbf{F}_1 + \mathbf{F}_{2} + \mathbf{F_{3}}) \label{eq:Fj} \\[6pt]
\mathbf{F}_1 &= 
- f_1(\psi, \gamma) \, \alpha_Y \, 
\operatorname{sign}\!\big(\mathbf{v} \cdot \mathbf{e}_{1_j} \big) \,
z \, A \, \mathbf{e}_{1_j} \label{eq:F1} \\[6pt]
\mathbf{F}_2 &= 
- f_{23}(\psi, \gamma) \, \alpha_X(\gamma, \beta, M) \,
z \, A \, \mathbf{e}_{2_j} \label{eq:F2} \\[6pt]
\mathbf{F}_3 &= 
f_{23}(\psi, \gamma) \, \alpha_Z(\gamma, \beta, M) \,
z \, A \, \mathbf{E}_3 \label{eq:F3}
\end{align}

where $\alpha_X$, $\alpha_Y$, and $\alpha_Z$ are resistive coefficients that represent stresses per unit depth, $\psi$ is the angle of $e_2$ relative to the horizontal velocity projection, $\gamma$ is the intrusion angle, $\mathbf{v}$ is the velocity, $z$ is the depth of the plate, and $A$ is the area of the plate \cite{treers_granular_2021}. \textit{M} represents a matrix used for fitting from Li et al \cite{li_terradynamics_2013}. $\alpha_X$ and $\alpha_Z$ are empirically derived by Li et al. and are functions of characteristic angles $\gamma$ and $\beta$. $f_1$, $f_{23}$, and $\alpha_Y$ are derived by Treers et al. \cite{treers_granular_2021}. $f_1$ and $f_{23}$ are scaling factors that are empirically characterized via horizontal plate drag experiments. Of the models for $\alpha_Y$, we choose the simpler implementation which assumes $\alpha_Y = \alpha_x(\gamma=0^\circ, \beta=0^\circ, M)$. We utilize the RFT coefficient, $\zeta$, to represent effective intrusion resistance from sand. We also employ the leading edge hypothesis \cite{askari_intrusion_2016} and remove forces on elements not pushing on the granular media, i.e. for which the normal dotted with the velocity is negative (Equation~\ref{eq:leh}). 
\begin{align}
\mathbf{v}_j \cdot \mathbf{n}_j < 0 \implies \mathbf{F}_j = 0 \label{eq:leh}
\end{align}
Using this formulation, along with the geometry of an intruding body, the overall reaction force from granular media is found by summing force contributions over all elements $(\mathbf{F}_{total} = \sum\textbf{F}_j)$. 

The matrix of generated forces then passes through single exponential moving average smoothing with an $\kappa_F$ (smoothing factor) of 0.5. $\kappa_F$ was tuned empirically to be 0.5: at this value, the simulated behavior at low velocities most closely matched experiments.
Tuning $\kappa_F$ results in force application magnitude changes depending on intrusion speeds (for more information, see Figure \ref{fig:model comparison}). With a smoothing factor of 0.5, the velocities decay exponentially to less than $0.1$~cm/s when the robot leg is in stance for 1.5~s. 
After the first timestep, the overall sorted force matrix (sorted form of Equation~\ref{eq:Fj} to match site and face index order) is smoothed to remove large spikes caused by sudden movements by the intruding body (Equation~\ref{eq:piecewise}).  
\begin{equation}
\mathbf{F}_{s_n} =
\begin{cases}
    \kappa_F \mathbf{F}_{s_n} + (1-\kappa_F)\mathbf{F}_{s_{n-1}}, & \exists \mathbf{F}_{s_{n-1}} \\[6pt] 
    \mathbf{F}_{O}, & \nexists \mathbf{F}_{s_{n-1}}
\end{cases} 
\label{eq:piecewise}
\end{equation} 
where $\kappa_F$ is the smoothing parameter, $\mathbf{F}_{s_n}$ is the smoothed and sorted force matrix at timestep $n$, $\mathbf{F}_{s_{n-1}}$ is the previous smoothed and sorted force matrix and $\mathbf{F}_{O}$ is the initial sorted force matrix. The smoothing allows the system to retain a small force that inhibits the velocity magnitude from reducing to zero. This prevents the system from experiencing unrealistic behaviors at low velocities, where force directions are not well defined.  As expected, without smoothing, end effectors oscillate vertically and horizontally with increasing amplitude. With too much smoothing, the intruders sink too deeply before resistance is encountered. 

For rotating intruder simulations, we used Proportional-Integral (PI) control for the integrated velocity motor from MuJoCo, with minor smoothing to achieve reliable motion (Equation~\ref{eq:CR} and Equation~\ref{eq:Cs}).  
\begin{gather}
C_R = C + C_{PI} + d \label{eq:CR} \\
C_{s_n} = \mu C_R + (1-\mu) C_{s_{n-1}} \label{eq:Cs}
\end{gather}
$C_R$ is the control signal from a predefined array of joint angles, $C$ is the velocity control value, $C_{PI}$ is the PI compensation, $d$ is a damping term, $C_s$ is the smoothed control signal, and $\mu$ is the smoothing parameter.
The PI gains were determined using trial and error. $\mu$ was set to 0.1 to minimally smooth the resultant control signal. Here, we use single exponential smoothing, similar to that in Equation~\ref{eq:piecewise}. Additionally, $\kappa_F$ was set to 0.1 for translating rotating intruder simulations to offset additional oscillations during excessive pivoting of the intruder.

To determine surface orientation, surface area, depth, and velocity direction in RFT-SiM, an RFT force calculation function takes built-in MuJoCo variables (mjData state variables) as variable arguments. Then, the function decomposes the velocity and normal vectors in the same fashion as described in Li et al. \cite{li_terradynamics_2013}, seen in Figure~\ref{fig:combo}. The MuJoCo API (application programming interface) state variable used for recording velocity between time-steps is the COM velocity for all bodies dependent on one another. For each body independent of one another, we instead use MuJoCo position state variables with kinematic equations to find the velocity of an intruder. 

When the program detects that part of a body is below the sand, it passes the necessary variables to the 3D RFT function. After the forces are calculated, the faces, sites, and forces are sorted by index to ensure proper placement of applied forces on the accompanying faces. To apply individualized forces, we index the 3-D force array and use \textit{mj.applyFT} from the MuJoCo API. Using \textit{mj.applyFT}, we apply forces at specific sites defined by the MuJoCo .XML. MuJoCo then uses the built-in physics solver to update multi-body dynamics and predict the motion of the system. Since there are many more force application points when using RFT-SiM, the total run time compared with the default MuJoCo contact model is an order of magnitude slower.  

\begin{figure}[ht]
    \centering
    \includegraphics[width=.8\columnwidth]{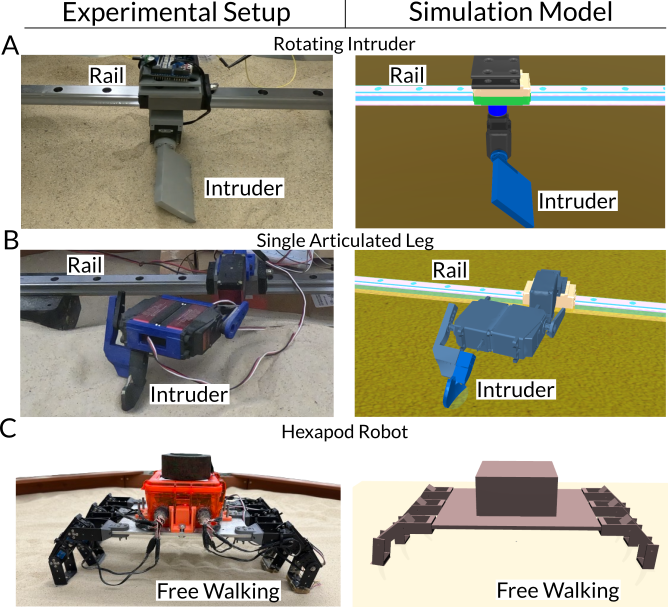}  %
    \caption{\textbf{We use three different experimental setups to explore and verify RFT-SiM.} (A) A rotating intruder is fixed to a rail to constrain motion to one direction parallel to the motor's rotation. This setup employs one actuator. (B) A 3-DOF single articulated leg is fixed to the sliding rail. This setup utilizes three actuators so that the leg can pull itself along the rail. (C) a freely-walking hexapod robot is placed in a sandbox. 
    }
    \vspace{-6mm}
    \label{fig:setup}
\end{figure}

\subsection{Simulation Setup} \label{Sim Setup}
The simulation initialization creates the bodies needed for RFT calculation, defines the necessary joints and sensor IDs to reference, and places sites on meshes in the descriptor. It is important to have even, dense meshing of the intruder in the force application region (as seen in Figure~\ref{fig:combo}) to ensure smooth force application across the body and limit inaccuracy caused by large plate sizes \cite{treers_granular_2021}. The model contains an equal number of force application sites to plates so that the program attributes a force to each discrete face \cite{angelidis_gazebo_2022}. 

Discretizing a body's faces into plates (meshing) is done before the simulation to prepare it to be read by \textsc{Open3D}, a library for 3D data processing \cite{zhou_open3d_2018}. 
To produce a mesh with a specified face count and a uniform distribution of faces, we import a CAD file into Fusion360. In Fusion360, we use the reduce and remesh tools to modify the mesh density. We find that a mesh density of 0.03 plates/mm$^2$, as shown in Figure~\ref{fig:combo}, provides adequate resolution for debugging and visualization. A denser mesh will result in longer, more accurate simulations; however, the benefits of a denser mesh decay when the local radius exceeds $\approx$5x the element characteristic length, as outlined in Treers et al. \cite{treers_granular_2021}. Using \textsc{Open3D} in RFT-SiM, the vertices and faces of the mesh are recorded. After collecting a dictionary of model bodies, the simulation has completed initialization and proceeds to the iterative simulation loop. The simulation timestep was chosen such that each simulation would produce stable results (approximately 0.0010 seconds for the rail-based simulations and 0.0008 seconds per iteration for the full robot).

We tested multiple intruder geometries, an articulated robotic scoop-like end effector, 
and a full 12-DOF hexapod robot in sand (shown in Figure \ref{fig:setup}). A descriptor .XML file for each of the testing environments was defined to closely match each corresponding physical testing environment. For visualization, we color the force application sites according to their representative stresses (Figure~\ref{fig:combo}).

\vspace{-3mm}

\subsection{Hardware and Software Programs} We utilized a custom built PC with an AMD Ryzen 7 9800X3D, 32 GB DDR4 RAM, and an NVIDIA GeForce RTX 5060 with Windows 11 for the simulations throughout this work. All simulations were conducted using the Python programming language. The open-source physics application Tracker was used to collect data from videos captured of experimental trials. The sensor used for force data during rotating intruder experiments was the BOTA MiniONE.

\vspace{-2mm}

\section{Results}
We established three baseline tests to verify the simulation method. The first baseline test measured force behavior as a result of rotating an object through sand (Figure \ref{fig:setup}A). This verified that our RFT model correctly predicted changes in force on an intruder as depth varied. The second baseline test measured the distance traveled by a carriage that was propelled along a rail by a rotating intruder  (Figure \ref{fig:setup}A). This test verified our model's ability to predict motion of a multi-DOF system. Third, we tested an articulated leg following a 3D path through sand to pull the carriage along the rail  (Figure \ref{fig:setup}B). This test validated that the model was capable of correctly integrating force predictions for 3D geometries and trajectories. Finally, we tested the full freely walking robot, without height constraints, to demonstrate that RFT-SiM can predict the motion of an unconstrained multi-DOF system  (Figure \ref{fig:setup}C). Simulations and physical trial examples can be seen in Supplementary Video 1. 


Different lab sand environments had different RFT coefficient $\zeta$ even though all used the same Quickrete play sand. Based on calibration tests outlined by Li et al. \cite{li_terradynamics_2013}, we vertically inserted a horizontal plate into the sand manually and measured the intrusion resistance of the substrate to determine $\zeta$ from the slope of force vs. depth. The sand in our wide, shallow (15~cm deep) sandbox has a measured RFT coefficient of $\zeta$=3.75 [N/cm$^3$], which is consistent with previous sand estimates for Quickrete play sand which typically show $\zeta$\textgreater1 for \cite{cao_mobility_2021}. This sand was used for constrained progress intruder, single articulated leg, and full hexapod robot experiments. Recently tilled Quickrete sand was used for the fixed-base intruder tests (Figure \ref{fig:torque}) in a deep plastic container (23~cm deep). Because the sand did not have weeks to settle, we measured $\zeta$=0.94 using the same characterization test. In simulation, we tested a range of RFT coefficients, from  $\zeta$ = 0.1 representing very loosely packed substrates, to $\zeta$\textgreater8 which is close to rigid flat ground (Figure \ref{fig:roboresults}CD, Figure \ref{fig:model comparison}A--D). 



\vspace{-2.7mm}
\subsection {Forces on Fixed-base Rotating Intruder}
\begin{figure}[ht]
    \centering
    \includegraphics[width=.8\columnwidth]{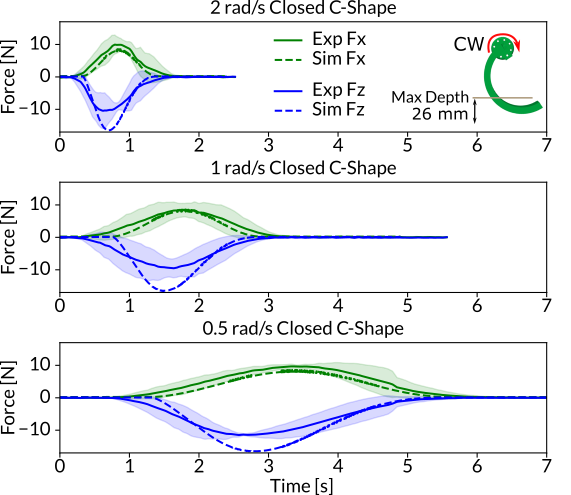}  %
    \caption{\textbf{
    When rotating a closed c-shape at constant angular velocity, RFT simulation correctly predicts increases in applied granular forces when the object is in sand.}
    (A) We verified the RFT implementation by rotating a c-shape $180^\circ$ about a fixed point such that the object enters and exits a flat bed of sand. 
    (B) During rotation, we measured the force horizontally (\textbf{x}) and vertically (\textbf{z}) (shaded regions represent $\pm$ one standard deviation about the mean over 5 trials).
   Differences in the shape and height of the peaks represent differences between experimental (solid line) sand and the RFT sand model (dashed line). Measured increased forces due to sand are within 15\% on average of RFT predictions while also varying rotational speeds. 
    }
    \label{fig:torque}
\end{figure}

\begin{figure*}[ht]
    \centering
    \includegraphics[width=.83\textwidth]{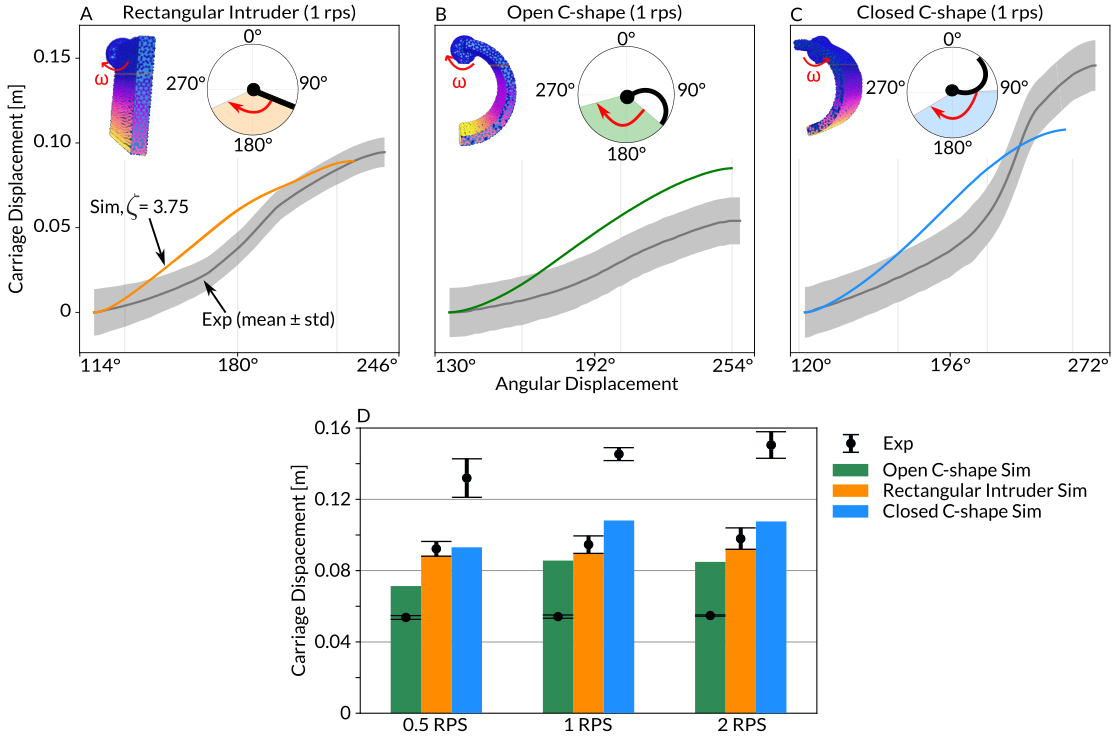}  %
    \caption{
    \textbf{When simple legs of different shape are rotated through the sand, RFT predicts motion along a fixed-height rail, with total displacements agreeing within  15\%.} (A) Carriage displacement vs. intruder rotation in both simulation and experiment is shown. Experimental data represents the mean and standard deviation over 5 trials. When the leg is a flat rectangular intruder, the carriage begins to move 
     as the intruder enters the sand and stops moving when the intruder exits, in both simulation and experimental trials. (B) When rotating a leg with an open C-shape through sand, the experimental carriage moves much less than the flat intruder leg, while the simulated carriage moves only slightly less. (C) When the C-shaped leg rotation direction is reversed (resulting in a closed C-shape), the carriage displaces more than for a flat rectangular intruder, and the simulation captures only some of this difference. Simulated forces on leading surfaces of each leg are shown in insets in (A-C). Circle insets show portions of rotation in which each leg engages sand. (D) When trials are compared across different speeds, the affects of the leg shape are present in both experimental and simulated trials, but the differences in performance are smaller in simulation. RPS represents radians per second.
     }
    \label{fig:flipresults}
    \vspace{-5mm}
\end{figure*}

The first baseline comparison between RFT-SiM and experiments is a 45~mm wide c-shape intruder that passes through a granular substrate by rotating about a fixed point (Figure~\ref{fig:torque}). Without RFT, MuJoCo cannot model passing through a granular substrate with existing rigid contacts: the forces increase indefinitely but the leg never passes through the substrate, which is modeled as a rigid body. When lowering solver impedance in MuJoCo, the solver has less power to enforce constraints, but still attempts to prevent the body from penetrating the surface (see Figure~\ref{fig:model comparison}). With RFT-SiM, the intruder is able to swing through the sand while the forces along the leg increase linearly with depth (see colors in Figure~\ref{fig:combo}D). 

For all rotating intruder tests, each experimental trial was run 5 times and position data was collected. The tested speeds were 0.5 radians per second, 1 radian per second, and 2 radians per second. These speeds were chosen to comply with the motor's speed limit of six radians per second and to stay within speed assumptions for simulation.
These tests were done at three speeds to confirm that resistive forces are independent of intrusion speed. For fixed-base rotating intruder tests, the experimental force was calculated from output readings from the BOTA Mini-One sensor. The Dynamixel, BOTA Mini-One, and intruder were suspended above the sand and attached to a carriage on a guide rail (Figure~\ref{fig:combo}).

The motor dynamics for the Dynamixel MX-64 used in the rotating intruder experimental setup were ignored and we assumed that the motor was capable of maintaining a constant velocity when commanded, to simplify the RFT force testing. For motor control of the rotating intruder system, we employed an integrated velocity motor as defined in MuJoCo for velocity control similar to that available for the Dynamixel MX-64. We measured the friction using an unloaded carriage slipping on a declined rail experimentally and then varied the coefficient of friction qualitatively to achieve similar starting and stopping of the carriage. The carriage was bisected laterally to provide horizontal planar contacts with the rail instead of using MuJoCo's convex hull. The convex hull assumes a convex approximation of a mesh body. If used, this would approximate the carriage as a box, leaving no inlet for the rail to pass as intended, and inhibiting motion along the rail. The intruder was discretized into plates that were evenly distributed across the body surface (Figure~\ref{fig:combo}) with a site and plate count of 5237 that ensured even, dense meshing with constant plate size.

With a c-shape attached to the rotating element, the simulation correctly predicts increases in force on the support beam that holds the object (Figure~\ref{fig:torque}). As the object rotates, the depth in sand increases, accompanying an increase in resultant force. Subsequently, the object rotates upwards, which decreases depth, and thus force. A slight delay in simulated force can be attributed to the physical experiment either having an uneven sand surface (causing an earlier interaction with the sand) or unexpected motor dynamics with the Dynamixel velocity controller. We do observe that simulation-predicted forces on the open c-shape exceeded those from the experiment. Additionally, the closed c-shape forces in simulation slightly underestimate the experimental forces. The BOTA MiniONE force sensor used was tared before each trial.

Across the different speeds tested, RFT-SiM predicts the same peak $F_z$ increase (16.2~N) and $F_x$ increase (8.5~N). This is consistent with the common RFT simplification at the heart of our implementation: that forces are independent of velocity magnitude.  At the fastest rotation, the maximum speed at the intruder tip is 18~cm/sec, which does not exceed 20~cm/s, a threshold above which inertial effects of the media become non-negligible \cite{agarwal_surprising_2021}. Nonetheless, at the fastest speed of 2 rad/sec (19 rpm), the magnitude of the force increase matches experimental data within a standard deviation. At slower speeds, there is a peak $F_z$ difference of 27\%. 

\subsection{Constrained Progress due to Rotating Intruder} \label{Constrained Prog}
Subsequent trials were conducted with the carriage allowed to freely translate along the rail. The translating trials used a flat rectangular 
intruder, an open c-shape, and a closed c-shape that were 3D printed with PLA filament (Figure~\ref{fig:flipresults}). These shapes were chosen to parallel RFT tests in Li et al. and Zhang et al. \cite{li_terradynamics_2013, zhang_effectiveness_2014}. Each intruder was placed such that the max intrusion depth was 5.5 centimeters, while attached to the rail 9 centimeters above the sand. 
The physical data for rotating intruders was collected by turning the intruder $180^\circ$ at a constant speed.

The carriage displacement after an intruder rotation was consistently within 6\% of the experimental value for the flat rectangular intruder (Figure~\ref{fig:flipresults}A). We observed a delay in the experimental carriage motion for each geometry (Figure~\ref{fig:flipresults}A-C) likely due to friction in the rail. The rail friction changed under different loads, and occasionally sand would enter the carriage bearings, also increasing friction. The total rail carriage displacement was highest for the closed c-shape. The closed c-shape had a convex geometry which was in contact with the substrate for the longest duration. The open c-shape resulted in less progress than the flat rectangular intruder in experimental data (-53\%), and to a lesser extent in simulation (-25\%). The effects of depth and angle are visualized in the inset of Figure~\ref{fig:flipresults}A-C. 
 
Our results show that the shape affects the experimental total progress more than the rotational speed.  Doubling and quadrupling the speed did not affect the experimental progress per cycle more than a standard deviation, which is expected given the velocity independence of the intrusion forces. The small variations in final displacement due to speed can likely be attributed to inertial effects of the rigid bodies. For the rectangular intruder and closed c-shape, the simulation predicted smaller changes in displacement with speed than the open c-shape. For the open c-shape, the simulation predicted a larger increase in displacement with speed than the flat rectangular intruder. The overall error roughly doubled between simulation and experiment for the fastest rotation speed relative to the lowest speed. However, flipping the shape of the leg from open to closed more than doubled the overall progress in experiment. These effects were more muted in simulation; however, the trend was consistent at all three speeds: the open shape resulted in less progress than a flat rectangular intruder which, in turn, progressed less than a closed shape. 
\begin{figure}[h!]
    \centering
    \includegraphics[width=.9\columnwidth]{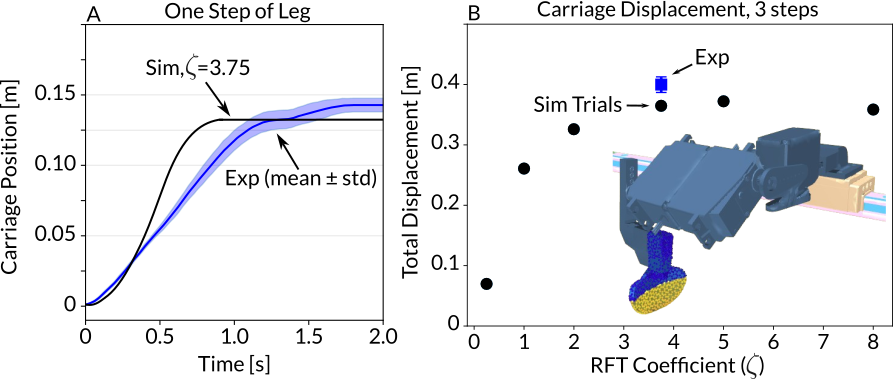}  %
    \caption{
    \textbf{The simulated articulated leg pulls the supporting carriage along the guide rail by using an end effector stepping into and out of sand.} (A) Time evolution of the guide rail carriage in simulation and experimental trials are shown. (B) Final carriage displacement after three steps of the articulated leg are shown at different simulated $\zeta$.
     }
    \label{fig:articulatedleg}
    \vspace{-5mm}
\end{figure}
\subsection{Constrained Progress due to Single Articulated Leg}

The third baseline test compared the carriage motion when propelled by a 3-DOF leg that swept through sand (Figure~\ref{fig:articulatedleg}). The articulated leg used the same carriage as the rotating intruder. The leg was attached to the carriage and allowed to pull itself by scraping along the top of the sand. The 3-DOF leg used three servo motors to trace a 3D path in and out of the sand. The motion of the end effector was comparable to an abducting step with a 3-DOF hexapod leg. The sand in these tests was placed 1~cm below the rail so the end effector would bury itself in the sand. The articulated leg system employs position-based controllers to model the chosen Savox servos and thus do not need smoothing. Plate count was set to 5237 for articulated leg simulations.

The articulated leg simulation predicted the displacement change over one step within 8\% of the experimental value (Figure~\ref{fig:articulatedleg}A). The simulated guide rail carriage position change (0.132~m) was less during each step than the experimental results (0.143~m), possibly due to the inertial effects, rail contact simplifications, or motor dynamics. 

\vspace{-3mm}
\subsection{Speed and Sinkage of Walking Hexapod Robot} \label{fullrobot}
\begin{figure*}[h]
    \centering
    \includegraphics[width=.82\textwidth]{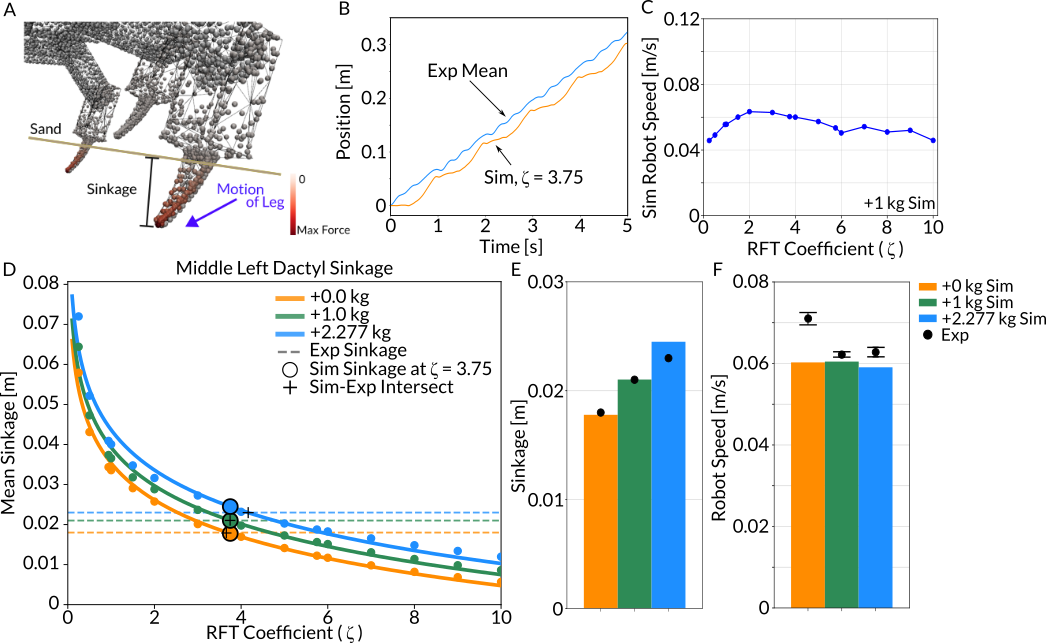}  %
    \caption{\textbf{We used our implementation to predict the motion of a walking hexapod robot.} (A) The legs and dactyl surfaces are discretized into plate elements, each with RFT ``sites'' (gray dots), with the deepest points generating the most force (graded in red). 
     (B) Translation of the robot without payload is tracked over five steps. 
     (C) The simulated robot speed is examined over $\zeta$ values ranging from 0.1-10. 
     (D) Mean sinkage vs. resistive coefficient for three different payloads is shown. The dactyl tips sink more when payload is added.  For each payload condition, experimental sinkages are averaged over each trial, and compared with simulation results across a range of RFT coefficients, $\zeta$. Based on initial sand characterization, we expected experimental $\zeta$ to be 3.75, which matches the results from experiment. The sinkage vs $\zeta$ was fit with a natural log function, $a * ln(b * x) + c$, where $a\approx-0.014$, $b\approx3.6$, and $c\approx0.056$. Robot sinkage (E) and speed (F) in RFT are compared with experiments across three payloads. Error bars represent one standard deviation.
     Experimental data in (F) is averaged over 5 trials $\pm$ one standard deviation while (E) is measured after a single trial.}
    \label{fig:roboresults}
    \vspace{-5mm}
\end{figure*}



To demonstrate prediction of motions for a 3D multi-body system, we evaluated the walking speed and penetration depth of the dactyls (sinkage) for a 4.04~kg 12-DOF hexapod robot (Figure~\ref{fig:roboresults}A) freely walking in a 15~cm deep sandbox (Supplementary Movie 1) \cite{grezmak_probing_2024}. The robot had knee and ankle joints on each of the six legs. The robot's end effectors are shown in Figure~\ref{fig:roboresults}A.  We ran 35 individual physical trials of 5 cycles (a cycle is a step by the robot) per trial (1 cycle per second), and 21 simulated trials of the same behavior (computation time of 2.1 minutes per trial on a desktop computer). For the full robot simulation, the resolution of each intruding body was set to 500 sites and plates to increase simulation efficiency compared to previous scenarios. The same gait and motor position commands were used on both the physical system and the simulation. We also varied payload by adding barbell weights weighing 0.5~kg, 0.75~kg, 1~kg, 1.5~kg, 2~kg, and 2.277~kg. The videos were taken by a camera placed above the sandbox at a distance such that that the camera distortion from the center to the edge of the sandbox was negligible. 

\begin{table}[b!]
\vspace{-20pt}
\centering
\caption{Clearance between Chassis and Sand while Walking (+0~kg)}
\vspace{-5pt}
\begin{tabular}{| c | c | c | c |} 
 \hline
 Sim Init [cm] & Sim Avg [cm] & Sim Final [cm] & Exp Final [cm] \\

 \hline
\textbf{5} & 12.1 & 12.6 &\\ 
\textbf{10} &  12.2 & 12.6 & $11.9 \substack{+0.7 \\ -0.4}$ \\
\textbf{15} & 12.3 & 12.6 &\\
 \hline
\end{tabular}

\label{table_chassis}
\end{table}

In experiment, the user placed the robot in the sand, whereas in simulation, the robot was dropped from 10~cm to allow the legs to move to initial position that matched experimental trials without lasting effects on the simulation. The drop height was determined by initializing the robot at a position where, after one timestep, the legs would transition instantly to the walking gait without getting caught too deep in the sand to walk. Simulation comparison with the experimental final sinkage for no-payload trials shows that the initial robot height in simulation does not affect the simulation robot height after 5 steps and varied minimally on average (Table \ref{table_chassis}).
The robot could have been initialized with submerged limbs, but RFT is dependent on velocity direction, so an initialization within the medium could cause unexpected behaviors. On the other hand, the MuJoCo physics solver produces infinite values when the robot is dropped from a large height (\textgreater10 [m]) into the RFT sand, so it is best to drop the robot from a reasonable height (0-1[m]) to avoid unexpected dynamics.

As shown in Figure~\ref{fig:roboresults}B, we compared the walking distance over time in both the experiment and simulation. Overall, RFT-SiM (with $\zeta$ = 3.75) tended to underestimate the walking distance per step between 3.4\% and 28\%. The estimated 5 step position error for no payload was 6.84\%. There was a visible rocking motion in each walking cycle of the simulated robot, which was not present in the physical trials.  In the example shown in Figure~\ref{fig:roboresults}B, the rocking motion occurs at each step near the end. In the physical trials, sand tends to pile up behind the legs (referred to as clumping \cite{kadanoff_built_1999}) which may steady the robot in transitions between alternating tripods of the walking gait. RFT-SiM does not capture this behavior because the substrate is assumed to be perpetually flat. An algorithm to estimate changes in penetration depth due to local surface deformation might mitigate rocking in simulation \cite{yu_modeling_2024}. 

As expected, in both the simulation and experimental data, the dactyls sink deeply into the sand, and this sinkage increases when additional payload weight is added to the robot (Figure~\ref{fig:roboresults}E). However, the simulation overestimates the sinkage depth of a dactyl (mean sinkage across three payloads is 2.6\% higher), which could be responsible for the lower speed in simulation (Figure~\ref{fig:roboresults}F), although rocking remains the main contributor.

To determine if tuning a single simulation parameter could be used to more closely fit the experimental speed and sinkage, we varied the RFT coefficient, $\zeta$. Although the calibrated experimental $\zeta$ is 3.75 in the sandbox \cite{li_terradynamics_2013}, increasing $\zeta$ in simulation is a stable way of decreasing sinkage depth. 
We also observe that the robot successfully walks when $\zeta$ is as low as 0.5, representing soft sand. When $\zeta$ is below 0.5, the robot fails to overcome slip due to to high sinkage and cannot move forward. $\zeta$ can be tuned depending on which quantities (speed, sinkage) are prioritized for accuracy.

\begin{figure}[h]
    \centering
    \includegraphics[width=\columnwidth]{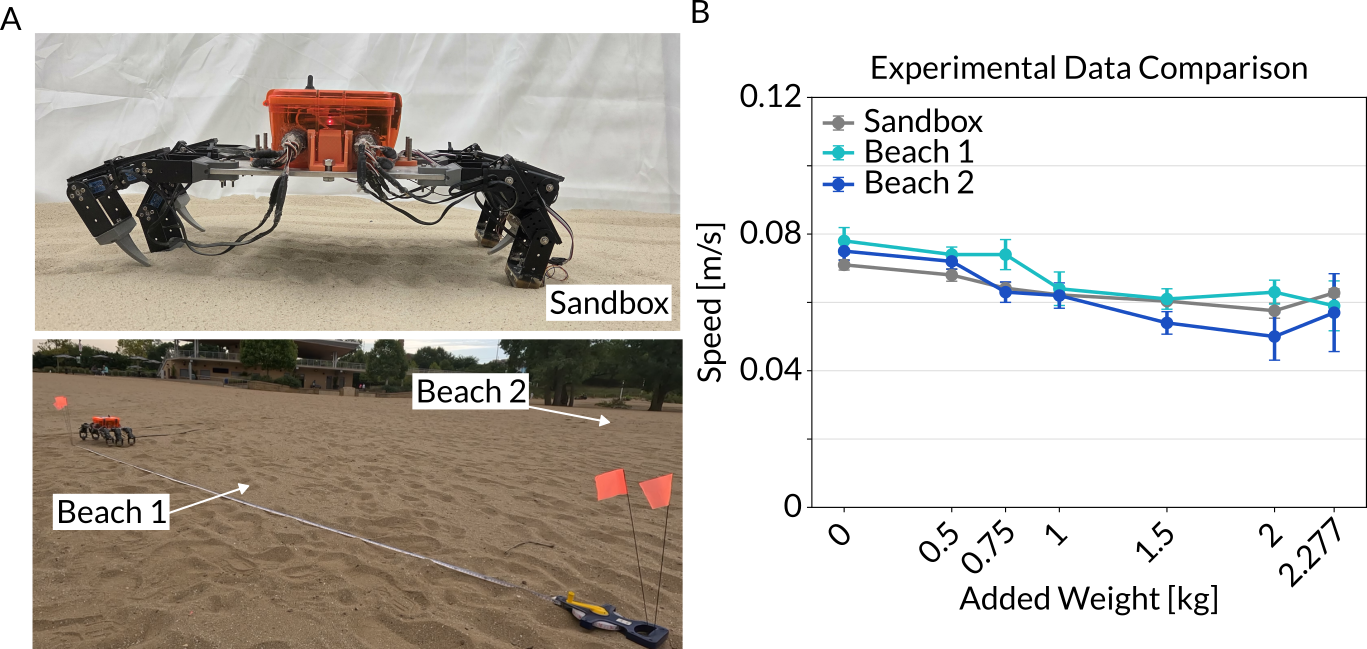}  %
    \caption{\textbf{The experimental sandbox data does not meaningfully differ from the uncontrolled beach sand.} (A) We tested walking speed in a lab sandbox and two different areas of a beach. (B) Each point and error bars represents averaging 5 walking trials and standard deviation.
    }
    \label{fig:beach}
    \vspace{-5mm}
\end{figure}

Increasing the payload affects the locomotion speed in nonlinear ways for both our simulation framework and for physical experiments. To confirm the simulation viability in varied sand areas, we compared the experimental results from the sandbox to two different beach environments (Figure \ref{fig:beach}A). Beach 1 was located closer to the shore while Beach 2 was farther from shore. Despite having $\zeta$ values of 0.96 for Beach 1 and 0.92 for Beach 2, the experimental walking speed remains close to the sandbox ($\zeta$=3.75) (Figure \ref{fig:beach}B). This relationship (Figure 7C) shows that in our $\zeta$ value regime ($\sim$1-5), the speed is relatively unaffected by the substrate $\zeta$ for this particular robot, but the sinkage still relies heavily on $\zeta$ (Figure 7D). The simulation predicted walking speed initially just for the sandbox, but this prediction can be applied to more real-world scenarios such as beach walking. The same trend was present in all experimental and simulation data; too much weight causes the robot to sink more deeply into the sand such that speed decreases.

\subsection{Comparison with other Contact Models}
\begin{figure*}[h]
    \centering
    \includegraphics[width=.9\textwidth]{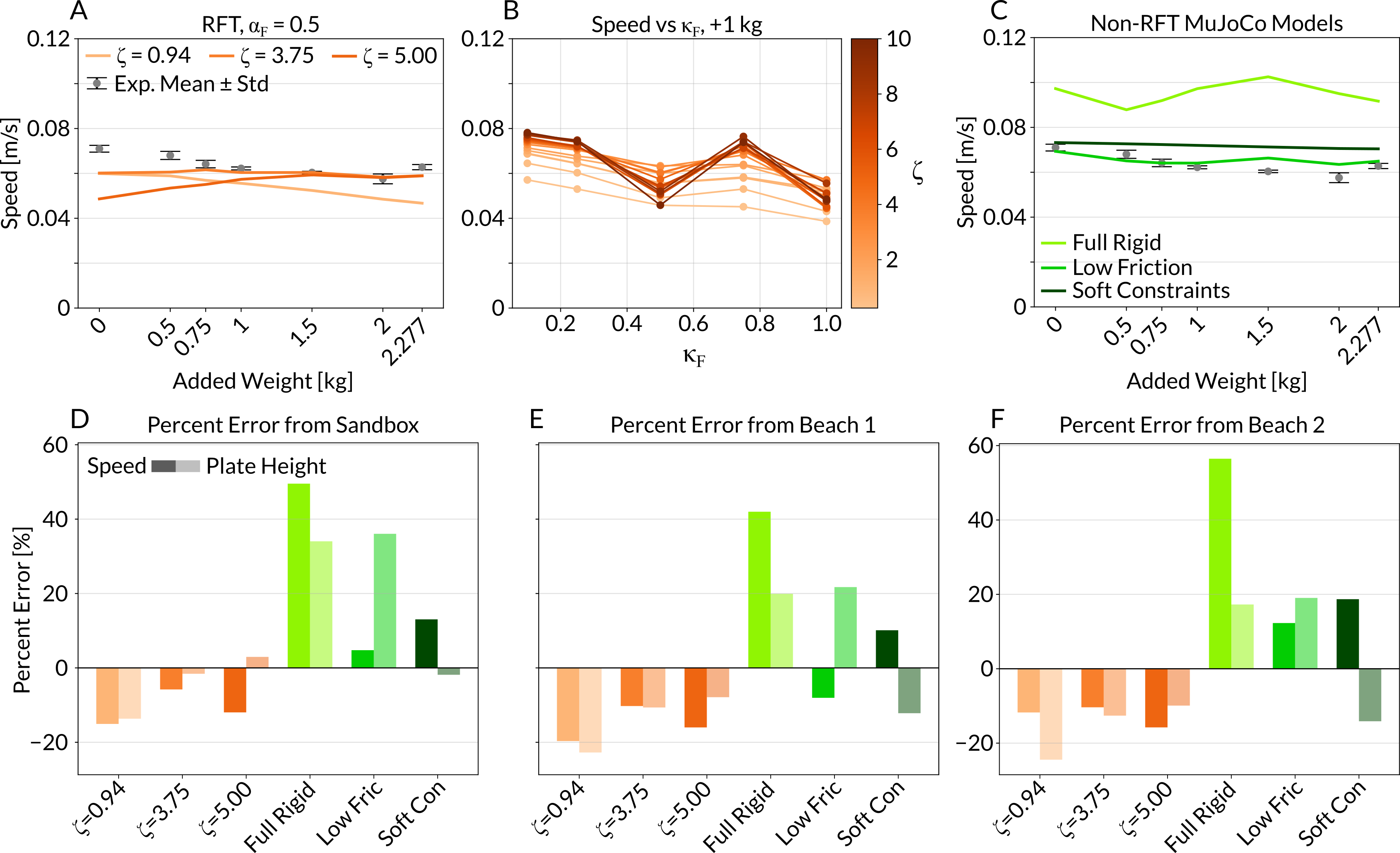}  %
    \caption{\textbf{RFT ($\zeta$ = 3.75) captures the mean speed and height better than existing options in MuJoCo while RFT ($\zeta$ = 0.94) models the payload effects of the experimental results.} We plot average walking speed vs. payload for six different simulation conditions, and compare experimental and simulation results. Error bars represent one standard deviation.
    (A) We simulated  payload increments with $\zeta$ = 0.94 (similar to beach sand), $\zeta$ = 3.75 (similar to sandbox), and $\zeta$ = 5.00. The sandbox data is plotted with error bars for comparison to the RFT simulations. 
    (B) The behavior in RFT speed estimation as $\kappa_F$ increases shows an inversion at $\kappa_F$ = 0.5 and non linearity when approaching 1. $\kappa_F$ = 1 represents no smoothing while $\kappa_F$ = 0 represents only smoothing (forces remaining at 0 with no regard for future values).
    (C) The fully rigid condition has a friction coefficient of 0.4. A low friction model is implemented with a friction coefficient of 0.008. Low Friction achieved the lowest speed error of MuJoCo methods. Soft MuJoCo constraints were obtained by modifying default impedance ($d$) parameters.
    (D-F) Mean error between each RFT/MuJoCo model and experimental results, for both horizontal speed (dark orange/green) and sinkage (light orange/green). With $\zeta$ = 3.75, matching the experimental sandbox, RFT achieved the lowest net error (when averaging speed and height errors).
    Each of the alternative simulation methods are labeled and contained within Supplementary Movie 1.
    }
    \label{fig:model comparison}
    \vspace{-5mm}
\end{figure*}

Other readily available contact models can be fitted to experimental results by tuning parameters, as evidenced in Figure \ref{fig:model comparison}C, but because of this required tuning they cannot directly predict walking speed a priori. To visualize this, we compared simulation data (Figure~\ref{fig:model comparison}C-F). Models chosen for comparison were: 1. MuJoCo soft constraints (low impedance in the solver), 2. Low Friction (rigid ground with noticeable slip $\mu$ = 0.008), and 3. Fully rigid (rigid ground with $\mu$ = 0.4). The models were chosen because they have easily tunable properties associated with ground contacts in MuJoCo. Each model was tuned to exhibit smooth motion over 5 cycles.
 
MuJoCo's soft constraint model (Figure~\ref{fig:model comparison}C) provides visually similar motions to experimental data. 
Tuning the soft constraint model involved changing the solver impedance for ground contacts, which corresponds to the simulated ground's ability to generate force. We used the built-in solver adjustments \textit{solimp} and \textit{solref} to modify contact dynamics for the soft constraint model. \textit{Solimp} adjusts the solver's impedance, or the ability of contacts to generate an opposing force \cite{todorov_mujoco_2012}. By default, \textit{solimp} is defined by 5 numbers that represent the impedance function $d(r)$ where $r$ is the constraint violation. We adjusted the first three parameters -- initial impedance $d_0$, impedance at width $d_w$, and width $w$ -- from $(0.9,~0.95,~0.001)$ to $(0.01,~0.5,~0.08)$ to weaken contact constraints for the Soft Constraints model (Figure~\ref{fig:model comparison}C). These weakened constraints allowed bodies to deeply intrude during contacts. To capture visually smooth motion during the Low Friction simulation (Figure~\ref{fig:model comparison}C), we modified \textit{solref} from $(0.02,~1)$ to $(0.02,~2)$, doubling the effective damping of body contacts. At low impedance, the end effectors penetrate the ground. However, because the constraint forces due to penetrating the ground are inaccurate, the model overestimates the speed. Furthermore, without accurate tangential resistance, all payload effects are erased as the speed decreases monotonically. 

Friction models inherently do not permit penetration, so high friction models (Figure~\ref{fig:model comparison}C) do not capture backward slipping motion and thus overestimate speed in sand, although the payload does affect speed. Lowering the friction to $\mu$ = 0.008, as in the Low Friction case (Figure~\ref{fig:model comparison}C), reduced the overall speeds to best match experimental data within 2.94~mm/s (4.76\%) across payloads. 
A fully rigid model is provided for comparison, but it did not match experimental speeds. For this model, the default impedance was used to represent a baseline walking simulation.

Increasing the payload in these models affected the robot's walking speed either irregularly (full rigid), with a slight increase (low friction), or decreasing monotonically (soft constraints). The predicted sinkage also suffers in these models. The full rigid and low friction models do not allow for dactyl penetration while the soft constraint is tunable to the correct sinkage, but not able to predict it. The simulation at $\zeta$=3.75 has a lower percent error average between height and speed from the Sandbox (3.7\%), Beach 1 (10.5\%), and Beach 2 (11.5\%) than all other methods tested (Figure \ref{fig:model comparison}D-F). Other methods, such as low friction, match either speed or height, but not both, after qualitative adjustments to align with data. RFT-SiM proves most viable when considering both speed and sinkage of the robot.

\vspace{-5mm}
\section{Discussion}
Together, these experiments show the utility of RFT-SiM for walking robots with varying penetration in sandy environments.
The accuracy is consistent with previous RFT implementations, within a framework (MuJoCo) that models multi-jointed robot and animal motion. As first approximations, the results across the tested systems show agreement within 34\%, capturing measurable trends in leg speed (Figure~\ref{fig:torque}), foot shape (Figure~\ref{fig:flipresults}), step progress (Figure~\ref{fig:articulatedleg}), sinkage (Figure~\ref{fig:roboresults}), and payload (Figure~\ref{fig:model comparison}). 

Compared with lab sandbox testing, RFT-SiM is fast, consistent, and economical, making it an attractive option for design iteration and early validation before lab and field testing. We were pleased that within two minutes, RFT-SiM delivers results with within 7\% error for 3D, unconstrained walking and sinkage of a robot (Figure \ref{fig:model comparison}D).  Other specialized granular media physics engines using a continuum approach (SPH) such as Chrono have run times of 1-2 hours for 3 seconds of simulation \cite{fang_chronogpu_2021, kozubek_chrono_2016}. In comparison, DEM models take from 1.2 minutes to 19 days on a high-powered multi-threaded GPU to achieve 5\% error for similar granular environments \cite{dosta_comparing_2024}.

We expect RFT-SiM to be useful in increasingly complex scenarios, with some inherent caveats.  One of the main simplifications of RFT compared to DEM is eliminating substrate particles from modeled state variables. 
Thus, history effects (e.g. from accumulated piles of scraped sand, trapped particles, or previous footprints) are not captured. Furthermore, forces are applied uniformly, neglecting edge flow, stick-slip friction, and jamming effects.


We expect that actuator modeling is a limiting factor for our experiments. Our experimental data on the carriage (Figures 5 and 6) often lags behind simulation predictions at the beginning of the step and then keeps moving after the step completes, suggestive of an inertial effect.  Similarly, for the freely walking robot, the simulation predicts larger variation in speed within a step cycle which is damped out in experimental data (Figure 7B). 
Because most of the weight of our robot is in actuators, the internal inertia and friction within the servomotors are likely important, but are currently unmodeled \cite{perez_inturias_virtual_2024,zakka_mujoco_2025}. However, because our robot's servomotors do not have limited position and torque feedback to inform motor model parameters \cite{kenneally_actuator_2020}, we wanted to keep motor models simple. 


Furthermore, continued refinements to increase accuracy are possible within this framework.  Newer RFT refinements have been developed, including inertial corrections \cite{huang_dynamic_2022, askari_intrusion_2016}, modifications for muddy or nonlinear soils \cite{kamrin_advances_2024,zhang_fluid-driven_2022,baumgarten_general_2019}, and adjustments for cohesive substrates \cite{kerimoglu_extending_2025}. Accurate inertial and sand buildup \cite{yu_modeling_2024} modeling would improve speed estimates as the robot moves through the sand (Figure \ref{fig:roboresults}BF). Sinkage estimations depends on the substrate interaction as the dactyl penetrates where cohesive, nonlinear, or muddy soil modeling would be beneficial (Figure \ref{fig:roboresults}DE). These modifications may be necessary to accurately explore the relationship between payload and speed, as early experimental and simulation data suggested that there is an intermediate payload where speed is optimal. However, more precisely calibrated sand beds \cite{menon_particle_1997} or comparison with calibrated DEM models \cite{coetzee_calibration_2009,hu_novel_2024} would be needed to validate these refinements. 

The two model parameters presented here represent different effects. The RFT coefficient, $\zeta$, \cite{li_terradynamics_2013} is meant to be constant for calibrated substrates (e.g. $\zeta$ = 3.75 for sandbox) and the pre-calibrated estimate has been reasonable across differently compacted beach regions. In general, we observe that walking speed is somewhat insensitive to $\zeta$ in the ranges 1 - 5 tested experimentally for our walking robot, but $\zeta$ plays a more significant role in accurately predicting sinkage. As a robot traverses new terrain, potentially with varying depths, packing, and grain sizes of granular media, adapting $\zeta$ would be a straightforward approach to maintaining model accuracy.   


We found a smoothing coefficient, $\kappa_F$, to be essential for adapting RFT to dynamic simulation.  MuJoCo relies on convergence over step-by-step approximations; while RFT relies on accurate motions of the intruding body. Without smoothing, rapid oscillations when a foot stops and starts may cause RFT force approximations to grow exponentially. With the drop initialization described above (Table \ref{table_chassis}), the effects of changing  $\kappa_F$ in the range of 0 to 1 are gradual (Figure \ref{fig:model comparison}B), which suggests a nominal constant value is appropriate.   Behaviors with high intrusion speeds or large downward forces may benefit from different smoothing approaches or refinements to the RFT model.  For example, drops from extreme heights could cause the robot to unrealistically ``bounce" without updated smoothing. Further studies may provide precise guides for choosing $\kappa_F$ in more complex simulations for which accuracy of below-surface dynamics is important. Recent results in elastic RFT or ``e-RFT'' could provide guidance for researchers seeking to maximize the accuracy of their simulations in low velocity regimes \cite{yilmaz_elastic_2026}.

\vspace{-4mm}

\section{Conclusion}
Our hope is that RFT-SiM will be valuable to others who are preparing robots for challenging outdoor and remote terrains, since running physics simulations for whole robot missions was previously computationally impossible in reasonable time frames \cite{govender_blaze-demgpu_2016, zhang_fluid-driven_2022, dosta_comparing_2024}. RFT-SiM is able to determine 12-DOF robot dynamics over 5 step cycles (one cycle per second) in about 2 minutes on a consumer-grade CPU. We have run RFT-SiM on an off-the-shelf 2019 laptop in 5 minutes for 5 cycles of a walking robot, suggesting potential for live field work simulations. Thus, researchers will be able to iterate quicker on robot hardware, controllers, and smart behaviors in a simulation “sandbox". RFT-SiM makes RFT-informed mechanical design (which has been done for low DOF robots \cite{zhang_mole-inspired_2025}) available for higher DOF robot bipeds, quadrupeds, and hexapods to learn to traverse and manipulate granular media. 
Users can visualize locations of maximum resistive force during a gait cycle, and vary a single parameter, $\zeta$, to represent different non-cohesive substrates.
Since the framework captures key trends in performance, future ML algorithms commonly used in MuJoCo \cite{he_learning_2024, xu_open-source_2024, fietkiewicz_neuromechanical_2025} could be used to systematically search the design parameter space to evolve robust new low-cost platforms for space exploration, desert operation, beach cleanup, and off-road mobility. 

We provide an open-source MuJoCo add-on called RFT-SiM that allows legged robots to walk and sink into granular media while still being computationally efficient and accurate. The code is currently available at [link redacted for double-blind review].

\vspace{-4mm}
\bibliographystyle{ieeetr}
\bibliography{SAnd}
\end{document}